\documentclass{ifacconf}

\usepackage{graphicx}      
\usepackage{natbib}        
\usepackage{amsmath,amssymb,amsfonts}
\usepackage{algorithmic}
\usepackage{graphicx}
\usepackage{textcomp}
\usepackage{xcolor}
\usepackage{url}
\begin{document}
\begin{frontmatter}

\title{Leveraging Deep Learning for 
Object and Position Recognition of 
Load Carriers for Autonomous Logistics Vehicles} 

\author[First]{Christoph Legat} 
\author[Second]{Tobias Miller} 
\author[Second]{Marco Riess}

\address[First]{Research Group on Cognitive Autonomy \& Predictive Intelligence, Technical University of Applied Sciences, Augsburg, Germany; \\ORCID: 0000-0002-5843-1845}
\address[Second]{Grenzebach Maschinenbau GmbH, Asbach-Bäumenheim, Germany; \\Email: \{tobias.miller,marco.riess\}@grenzebach.com}

\begin{abstract}                
This work explores the use of artificial intelligence in mobile robotics to achieve autonomous detection and pose estimation of load carriers for automated pickup. A deep neural network is designed to recognize predefined landmarks on the carrier from RGBD data; these landmarks are then used to compute the carrier’s pose. The network operates directly on RGBD images to estimate landmark positions, which form the basis for determining the carrier’s location. The approach is validated in extensive experiments and comprises both software and hardware implementations.

A deep learning–based framework is presented to detect load carriers and estimate their pose for use with autonomous logistics vehicles. Our method uses a convolutional neural network to identify characteristic reference points on the carrier from RGBD input and computes its pose by combining these inferred landmarks with prior geometric knowledge. Experiments show that the resulting accuracy is sufficient for reliable load carrier detection in industrial environments, confirming the suitability of the method for autonomous intralogistics applications.
\end{abstract}


\end{frontmatter}

%
%
\section{Introduction}
\label{sec:intro}
A large share of the value added by manufacturing firms arises from transforming externally supplied materials and components into products with higher economic worth. This value creation takes place within the companies’ production facilities, where the reliable availability of required materials at each processing step is essential. This requirement holds for both job-shop and flow-shop production, independent of the specific shop-floor organization. Intralogistics is therefore central to value creation, as it enables the efficient transfer of input materials and components between receiving, storage, and processing stations, thereby maintaining an uninterrupted material flow—a core intralogistics function that has been widely studied \citep{scholz_integrating_2016}. At the same time, rising global wage levels and persistent shortages of skilled labor pose additional challenges for manufacturers. In high-wage regions such as Europe, automation of intralogistics processes has become crucial for maintaining the competitiveness of production sites. Automation solutions typically implement automatic functions in accordance with IEC~\mbox{60050-351}~\citep{noauthor_din_2014}; in warehouses, for instance, conveyor systems enable automated goods transport. However, such static material-handling systems increasingly clash with the demand for flexible shop-floor organization required for highly variable, adaptable, and thus competitive manufacturing \citep{winter_evolutionary_2020, kraft_new_2021}. As a result, autonomous mobile vehicles are gaining importance for handling flexible material flows in production environments, since they can move freely within their surroundings and dynamically adapt to changing tasks. These mobile robots, hereafter termed autonomous guided vehicles (AGVs), depend on sensor-acquired and -processed data for self-directed navigation. Artificial intelligence offers substantial potential for autonomous driving by enabling automated obstacle detection and subsequent planning and execution of driving maneuvers. This thesis investigates the application of artificial intelligence, with an emphasis on deep learning, for robust detection of goods carriers in industrial settings. The central aim is to infer the AGV’s position and thereby provide the basis for an appropriate and optimized approach strategy.

This work presents an integrated framework for jointly recognizing load-carrier objects and estimating their poses in industrial environments. It proposes a deep learning–based method to identify characteristic load-carrier landmarks from RGB-D observations, followed by a lightweight geometric routine that computes the relative pose of the carrier based on its known 3D model. The approach is implemented as a real-time system on industrial-grade hardware. The proposed method is empirically evaluated in a industrial production setting, with a focus on execution speed, pose estimation accuracy, and robustness under varying operating conditions.

The remainder of the paper is structured as follows: Subsequent Sect.~\ref{sec:requirements} details the requirements for an approach to object and position recognition based on machine learning applicable in industrial systems. These findings form the basis for the analysis and dissemination of existing work in Sect.~\ref{sec:sota}. The developed approach is described on the conceptual level in Sect.~\ref{sec:concept}. The implementation and exhaustive evaluation of the approach in real industrial environments are presented in Sect.\ref{sec:eval} before the paper is concluded in Sect.~\ref{sec:summary}

%
%
\section{Use Case \& Requirements}
\label{sec:requirements}
\begin{figure}
    \centering
    \includegraphics[width=0.7\linewidth]{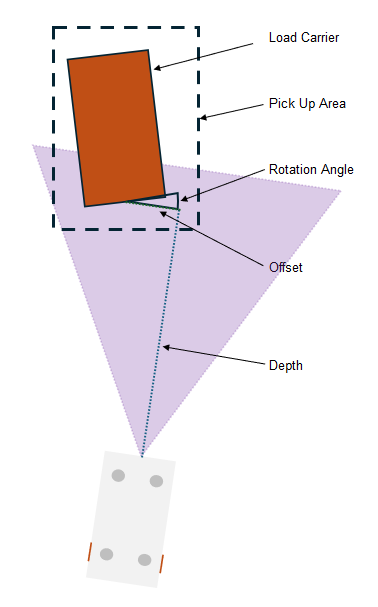}
    \caption{Exemplary visualization of the aspects of object detection and positioning (depth, offset and rotation angle)}
    \label{fig:use_case_visu}
\end{figure}

This study aims to detect load carriers and estimate their poses for integration with AGVs. Utilizing AGVs for transporting load carriers in production environments is an effective approach to intra-logistics \citep{fragapane_increasing_2022, scholz_integrating_2016}. In such applications, AGVs autonomously transfer load carriers between manufacturing and assembly stations. To this end, predefined intra-logistic areas (ILAs) function as intermediate buffer zones where load carriers are deposited and retrieved. Because human operators manually place the load carriers after processing, their arrangement within the ILAs is non-deterministic. In the industrial scenario considered here, a load carrier may exhibit a lateral displacement of up to ±20 cm and a rotational deviation of up to ±10° (see Figure \ref{fig:use_case_visu}). Consequently, the primary challenge for autonomous transport is the automatic acquisition of the load carrier. This requires robust detection of the object and its pose (depth, lateral offset, and orientation angle) with a maximum error of 1\% relative to the depth along the AGV’s approach path, ensuring reliable and collision-free pickup.

To apply object recognition and position in industrial settings, sufficient performance is required to avoid short stops. This resulting non-continuous movements can lead to, in addition to loss of performance, increased wear and tear on the hardware. As a consequence, the algorithmic performance of object detection and positioning must be adequately fast in terms of the sampling rate of the sensors used and the driving speed of the AGVs (requirement~\mbox{Req-1}).
%
Furthermore, the application requires object and position detection relative to the position of the AGV to facilitate movement in a continuous manner when approaching the receptacle for goods. Consequently, a suitable system requires determining the relative position of the load carrier with reference to the AGVs (Req-2). 
%
The system is used in industrial environments (mainly production facilities). When applying machine learning, it is evident that an approach is trained with industrial data because lighting conditions, surroundings, and backgrounds are typically very similar to each other but differ significantly from other environments. Consequently, training based on industrial environment data and testing within them is important to ensure reliable industrial applicability (Req-3).

%
%
\section{State of the Art}
\label{sec:sota}

The main requirement for the system is runtime, which is specified by the real-time capability of the depth camera with 15 fps. Looking at the works \citep{li_application_2019, muhammad_yasir_3d_2022, he_ffb6d_2021, wang_densefusion_2019}, they fulfill the run-time criteria. It is noticeable that the CNN network architecture is predominantly used. 
Position detection is a crucial aspect as the goal of the system, which is implemented in this work. There are already many different approaches to determining the position of an object. \citep{he_ffb6d_2021, wang_densefusion_2019, xiang_posecnn_2018, he_pvn3d_2020, ausserlechner_zs6d_2023} recognize points on objects and iteratively reduce the error between the recognized points and the geometry of the known object. The position of the object geometry with the minimum error corresponds to the real position. The necessary transformation of the object is used to calculate the rotation and translation. \citep{lin_hipose_2023} describes the surface of the object with a hierarchical binary code. The elements of the point cloud are assigned to the code, whereby the first digits of the code correspond to a rough positioning. The last digits are determined iteratively to refine the model. The position is determined from the links between the point cloud and the binary code. \citep{wen_foundationpose_2024} recognizes the object and compares it with known views to determine an initial position. This position is compared with the corresponding screen section and improved iteratively. \citep{lin_sam-6d_2024} performs a segmentation to predict poses with part of the data using a transformer. The best pose is selected and refined with all the data points to make a more accurate prediction. Another approach is described in \citep{xu_cxtrack_2023}. Here, the position is not recognized but tracked. An initial position is therefore required. The network compares the current with the previous point cloud to determine the changes in position. This change is applied to the previous position to determine the new one. 
The final differentiator is the use case. Most applications are created and tested using public benchmark datasets. These rarely represent industrial environments. Research work related to industrial environments is \citep{li_application_2019, hung_integrating_2022}. Here, pallets are recognized by means of a neural network. These are to be picked up by AGVs. Object recognition during movement is dealt with in \citep{lee_object_2021}. Obstacles in the paths are recognized. \citep{clavero_dmzoomnet_2024} improves object recognition by enlarging distant objects in order to achieve better recognition by the neural network. As illustrated in Table ~\ref{tab:papers_criteria}, the assessment of the existing approaches described above reveals their respective levels of fulfillment of requirements; none of them is able to fulfill the given industrial requirements.
In contrast to existing approaches, the proposed method focuses on the combination of lightweight landmark detection and geometries-based positioning specifically tailored to real-time constraints on industrial embedded hardware.

    \begin{table}[h]
        \centering
        \begin{tabular}{|c|c|c|c|}
            \hline
            \textbf{Papers} & \multicolumn{3}{c|}{\textbf{Requirement}} \\ \cline{2-4} 
            & \textbf{Req-1} & \textbf{Req-2 
            } & \textbf{Req-3 
            } \\ \hline
            \cite{muhammad_yasir_3d_2022} & \checkmark & $\times$ & $\times$ \\ \hline
            \cite{he_ffb6d_2021} & \checkmark & \checkmark & $\times$ \\ \hline
            \cite{wang_densefusion_2019} & \checkmark & \checkmark & $\times$ \\ \hline
            \cite{li_application_2019} & \checkmark & $\times$ & \checkmark \\ \hline
            \cite{xiang_posecnn_2018} & $\times$ & \checkmark & $\times$ \\ \hline
            \cite{lin_sam-6d_2024} & $\times$ & \checkmark & $\times$ \\ \hline
            \cite{he_pvn3d_2020} & $\times$ & \checkmark & $\times$ \\ \hline
            \cite{lin_hipose_2023} & $\times$ & \checkmark & $\times$ \\ \hline
            \cite{xu_cxtrack_2023} & $\times$ & \checkmark & $\times$ \\ \hline
            \cite{ausserlechner_zs6d_2023} & $\times$ & \checkmark & $\times$ \\ \hline
            \cite{wen_foundationpose_2024} & $\times$ & \checkmark & $\times$ \\ \hline
            \cite{hung_integrating_2022} & $\times$ & \checkmark & \checkmark \\ \hline
            \cite{lee_object_2021} & $\times$ & $\times$ & \checkmark \\ \hline
            \cite{clavero_dmzoomnet_2024} & $\times$ & $\times$ & \checkmark \\ \hline
            \cite{bohm_bildanalyse_2020} & $\times$ & $\times$ & \checkmark \\ \hline

        \end{tabular}
        \caption{Overview of requirement fulfillment for related works}
        \label{tab:papers_criteria}
    \end{table}

%
%
\section{Concept}
\label{sec:concept}
\begin{figure}[htb]
    \centering
    \includegraphics[width=1\linewidth]{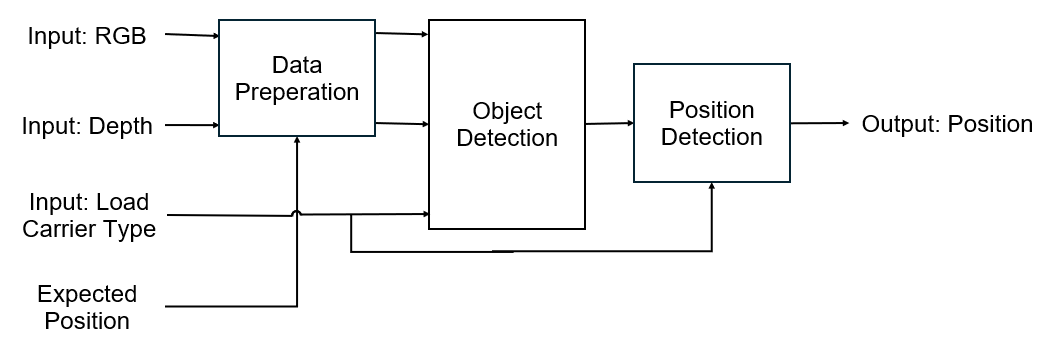}
    \caption{Proposed concept for object and position detection}
    \label{fig:concept_overview}
\end{figure}

The proposed concept is based on a three-stage approach, which depends on disparate input data. As illustrated in Figure~\ref{fig:concept_overview}, the process of preparing the image data (RGB) and the depth information is a prerequisite for subsequent analysis.
Afterwards, given the preprocessed RGBD input data, the task is to detect the load carrier and estimate its pose relative to the camera (object detection). We assume that the 3D model of the object, i.e. the load carrier, is available as is typically the case in intra-logictics industrial settings. The object detection is realized by means of landmarks detection using a CNN architecture. Afterwards, these predicted landmarks and the available 3D model of the load carrier are used to calculate the position (position detection). 

The position consists of depth, offset, and rotation angle, as introduced in Section~\ref{sec:requirements}. The AGVs know the rough position and the type of load carrier during the pick-up task. As soon as the AGVs are in the vicinity of the load carrier, the approach begins. First, it reads the RGBD data from the depth cameras. They are pre-processed for object recognition and transferred to a CNN. The region of interest is cut from the image to generate better starting conditions for the neural network. In addition, the background is masked out based on the depth data. In object recognition, both color and structural features are extracted and combined to make a prediction. The position is then calculated on the basis of the known geometry of the load carriers and the predicted landmarks.

\subsection{Object Detection}
\begin{figure*}[htb]
    \centering
    \includegraphics[width=.8\textwidth]{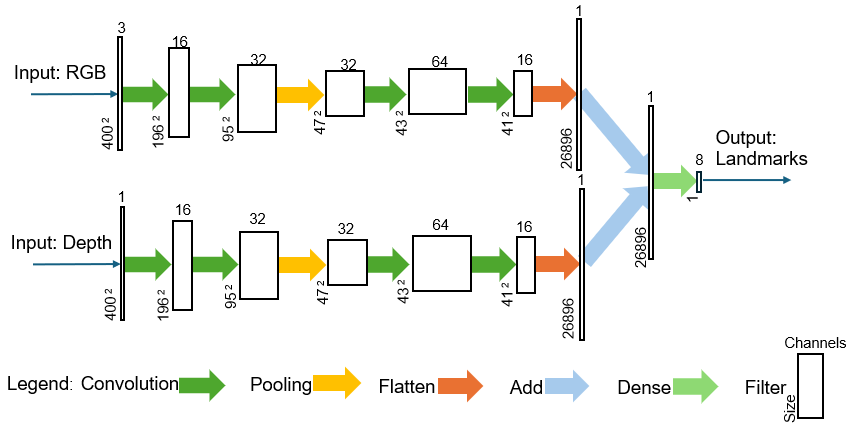}
    \caption{Neural Network Architecture for Landmark Detection of Load Carriers}
    \label{fig:neural_network_landmark_detection}
\end{figure*}

The task of object detection is to recognize the corner points of the charge carriers using a neural network. It receives the RGBD data recorded by the depth camera and the load carrier type as inputs. A CNN as seen in Figure~\ref{fig:neural_network_landmark_detection} is used as the architecture due to its fast interference time. 

The relevant features are extracted from the camera data through several convolution and pooling layers of the neural network and combined with each other. As in \citep{ronneberger_u-net_2015} and \citep{xiang_posecnn_2018}, the resolution is gradually reduced through the use of max-pooling layers, while the number of filters is increased to capture a deeper feature representation. The output is chosen to be landmarks instead of bounding boxes, in order to better understand the position of the load carrier in the picture. To output the landmarks, the feature representations of color and depth data are converted into one-dimensional vectors and combined. Fully connected layers are used to generate the eight output coordinates. These represent the landmarks with their X and Y coordinates in the camera data and are further processed by position detection.

\subsection{Positioning}
\begin{figure}[ht]
    \centering
    \includegraphics[width=1\linewidth]{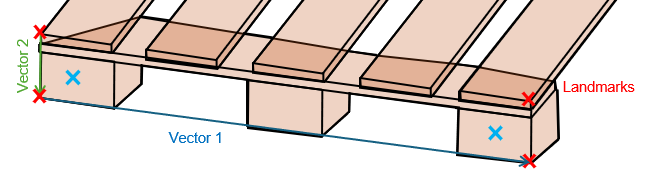}
    \caption{Calculation the depth reference points (light blue) using predicted landmarks (red) and vectors from the known object geometry}
    \label{fig:position_calculation_points}
\end{figure}

Position detection exploits the fact that load carriers in industrial environments are usually rectangular. Therefore, their corners serve as distinctive landmarks, as illustrated in Figure~\ref{fig:position_calculation_points}. Since the load carriers rest on the floor, position detection yields a 6D pose. It must estimate the distance between the camera and the load carrier, as well as lateral offset and rotation, which are subsequently compensated by additional motions of the automated guided vehicle during pick-up. To this end, position detection computes reference points on the load carrier of its type and predicted landmarks. The neighborhoods around these reference points are evaluated to suppress noise. The depth values in these regions are then used to derive the rotation angle and the distance to the load carrier. The horizontal offset is obtained from the deviation of the load carrier from the center of the image. The center point of the load carrier is inferred from the landmarks and the type of load carrier; its displacement from the image center, in pixels combined with its depth, defines the offset. The three quantities—depth, horizontal offset, and rotation angle—together describe the relative pose of the load carrier with respect to the depth camera.

For example, a detected load carrier in the RGBD data is fed into the neural network, which outputs the pixel positions of its four corner landmarks. From these landmarks, reference regions on the carrier are defined and depth values within them are averaged to attenuate sensor noise. Combining these averaged depths with the known carrier geometry enables us to estimate its rotation angle, distance, and lateral offset with respect to the AGV.

%

%
%
\section{Implementation \& Evaluation}
\label{sec:eval}
The evaluation is based on real-world measurements conducted in an industrial test setup, where load carriers were placed at varying distances, offsets, and rotation angles relative to the depth camera. The trained neural network was evaluated using the validation dataset and additional measurement series with controlled positioning variations.
\subsection{Implementation}
The approach is realized through a combination of software and hardware components. For capturing RGBD data, we utilize an Intel RealSense D455 depth camera\footnote{\url{https://www.intel.com/content/www/us/en/architecture-and-technology/realsense-overview.html}}. Due to its global shutter feature, the sensor records all its elements at once, minimizing distortion and noise in motion. The software is implemented in Python and is executed on an industrial PC. The camera data are read through the PyRealsense2 library\footnote{https://pypi.org/project/pyrealsense2/}. The neural network is developed with TensorFlow\footnote{https://www.tensorflow.org/}. For data annotation, an external tool is used. 

The software is structured into four categories: data preparation, training, application, and experiments. In data preparation, the recorded videos are converted into images and stored in the corresponding folder structure to enable external annotation. Training loads the annotated dataset and creates the neural network. The dataset used for training and validation consists of approximately 600 manually annotated RGBD frames recorded in an industrial environment using the Intel RealSense D455. The data capture realistic variability in lighting conditions, backgrounds, and positioning inaccuracies typical for production environments. The data set was divided into training and validation subsets.When training the model, the neural network is adapted and trained until the required accuracy is achieved. The model is then saved with its weights. After completion of the training, the model is saved. During application, the trained model is loaded, and the captured RGBD images are used to predict the landmarks. The corresponding image section is selected from the read-out RGBD-data and the predicted position and delivered to the neural network. This predicts the landmarks, which are then used to determine the position by recognizing the position. After that, the position is calculated. The experimental section is designed to implement tests for assessing the concept.


\subsection{Evaluation of Object Detection} 
\begin{figure}
    \centering
    \includegraphics[width=1\linewidth]{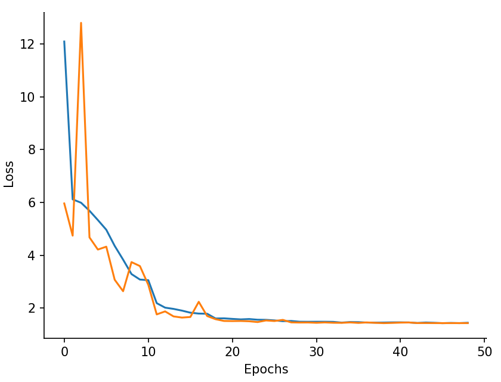}
    \caption{Training (blue) and validation (orange) loss of the neural network}
    \label{fig:loss_neural_network}
\end{figure}

The neural network is trained using an annotated data set manually. An implemented function is used to calculate the loss function to determine the error for each of the four corner points and scale it to the width of the load carrier. The mean value of the errors is returned. This provides information on the average accuracy of object recognition. After going through the training process as visualized in Figure~\ref{fig:loss_neural_network}, the neural network achieves a validation error of 1.4 after 20 epochs, which corresponds to an average absolute error of 1.4cm through the validation data set due to scaling.

The integration of color and depth data in the neural network surpasses the performance of networks that utilize each type of information individually. This validates the hypothesis of improved accuracy and highlights the necessity of incorporating both data types for maximal performance. The network architecture was established through iterative training and reflects the optimal configuration achieved.
The runtime of the system is evaluated after the training process using the validation dataset. If the data set is passed to the neural network, it achieves a run-time of up to 7.2ms per image. In case of a real-time capable application,the data recorded by the depth camera are transferred to the network picture by picture individually. This reduces the run-time to at least 34 ms because calling the network requires most of the run-time. However, the set requirements are still achievable.

\subsection{Evaluation of Positioning Accuracy}
\begin{figure}
    \centering
    \includegraphics[width=1\linewidth]{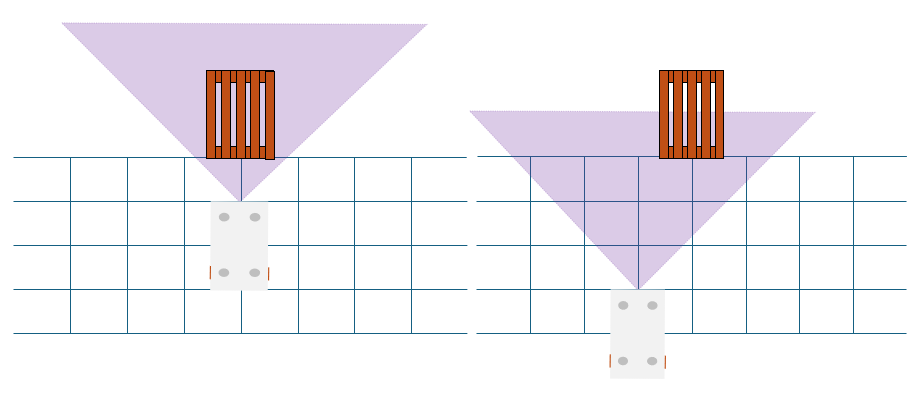}
    \caption{Evaluation of the position detection algorithm}
    \label{fig:eval_position_detection}
\end{figure}

To assess position detection accuracy, the loader is rigidly fixed as shown in Figure~\ref{fig:eval_position_detection}. The depth camera is calibrated to the load carrier using a checkerboard pattern, introducing variations in depth and lateral offset. Further measurement series are performed for different rotational orientations. With the load carrier aligned parallel to the depth camera, a maximum measurement distance of 1 m yields an accuracy of 1\%. Increasing both distance and rotation of the load carrier leads to higher position errors, while rotational deviation remains below 1° for all tests. The average computation time for position estimation over all measurements is 21.8 ms.

\subsection{Evaluation of the Overall Approach}
The load carrier is positioned at a fixed location relative to the depth camera, and random perturbations are added to this nominal pose to emulate misplacement. The perturbation magnitude is varied from zero to a maximum of 5 cm and 10 cm, respectively. In the case of correct placement, 80.6\% of the predicted pose elements lie within the acceptable tolerance; 71.4\% of the remaining incorrect elements are offset-related prediction errors. With an introduced pose uncertainty of up to 5 cm, 72.9\% of the predicted pose elements remain within tolerance, and all misclassifications correspond to offset errors. When the expected position is perturbed by up to 10 cm, the method degrades markedly: only 33\% of the predicted elements satisfy the tolerance constraint. Table~\ref{tab:Evaluation des Gesamtkonzeptes} presents the mean errors in depth, offset, and rotation angle across the different tolerance intervals. Larger deviations in the expected pose lead to increased offset errors due to incorrectly predicted landmarks. If the traces are no longer on the load carrier, the depth and rotation errors rise sharply.
%
\begin{table}[h]
    \centering
    \begin{tabular}{|c|c|c|c|}
        \hline
        \textbf{Positioning} & \multicolumn{3}{c|}{\textbf{Mean Error}} \\ \cline{2-4}
       \textbf{Tolerance} & \textbf{Depth [mm]} & \textbf{Offset [mm]} & \textbf{Angle [°]} \\ \hline
        0 cm & 4,22 & 12,64 & 0,53 \\ \hline
        0-5 cm & 2,38 & 15,47 & 0,41 \\ \hline
        0-10 cm & 70,10 & 60,60 & 9,36 \\ \hline
    \end{tabular}
    \caption{Evaluation results at different positioning tolerances}
    \label{tab:Evaluation des Gesamtkonzeptes}
\end{table}
\subsection{Discussion}
The proposed concept has not yet reached adequate precision to effectively detect the load carrier. The primary issue lies in predicting the landmarks with high accuracy. This can be enhanced by employing larger datasets for training and designing a more sophisticated neural network. Nevertheless, the present concept operates at a speed of 55.8 ms, which is just shy of the needed 66.6 ms runtime. It is apparent that the majority of the runtime is consumed in loading the neural network. Advancements in these aspects might allow the neural network to grow in complexity and increase its accuracy.
The use of landmarks to identify load carriers offers a promising method. Their stationary placement on the load carrier gives more data than bounding boxes, facilitating subsequent position determination. The similarity of neural networks used in landmark detection to those in bounding box prediction yields anticipated performance levels, making them especially suitable for real-time use.
The main limitation of the approach is the dependence on accurate landmark prediction. In cases of large deviations from the expected position, incorrect landmark predictions lead to significant positioning errors.

\section{Summary \& Outlook}
\label{sec:summary}
This paper provides a contribution to the automation of intra-logistic processes by means of automated recognizing load carriers and its positions (distance to AGV, rotation). The key challenge lies in precisely detecting the load carrier's position, necessitating algorithms that handle changes in position with less than 1\% deviation. 

We proposed a three-stage method for object detection, utilizing RGBD data and CNN architecture to recognize the landmarks of load carriers and estimate their positions accurately. The method integrates hardware and software and was implemented using an Intel RealSense depth camera and TensorFlow for neural network development. The system's precision and speed, although promising, require further improvement for optimal accuracy in real-time applications.

In a nutshell, while the current concept of using landmarks for load carrier identification shows promise, especially in real-time environments due to its speed, enhancing datasets and neural network architecture could further refine accuracy. Furthermore, the combined application of color and depth information showed a positive impact, improves the neural network's performance compared to using single data.

\bibliography{zotero}             

\end{document}